\documentclass[letterpaper, 10pt, journal, twoside]{./Style_template/RSS/IEEETrans}  

\usepackage[letterpaper, left=0.7in, right=0.7in, bottom=0.76in, top=0.7in] {geometry}
\def\IEEEtitletopspace{0.2in}
\IEEEoverridecommandlockouts                              
\usepackage{cite}

\usepackage{amsmath}
\usepackage{moreverb,url}
\usepackage{times}
\usepackage{multicol}
\usepackage{graphicx}
\usepackage{amssymb} 
\usepackage{gensymb}
\usepackage{amsmath, xparse}
\usepackage{mathrsfs}
\usepackage{soul}
\usepackage{enumitem,kantlipsum}
\usepackage{multirow}
\usepackage{xcolor}
\usepackage{physics}

\usepackage[colorlinks,bookmarksopen,bookmarksnumbered,citecolor=red,urlcolor=blue]{hyperref}

\newcommand\BibTeX{{\rmfamily B\kern-.05em \textsc{i\kern-.025em b}\kern-.08em
T\kern-.1667em\lower.7ex\hbox{E}\kern-.125emX}}

\begin{document}

\title{Terrain-Dependent Intra-Cycle Leg Timing for Effective Locomotion on Granular Slopes}

\author{Long Zhao$^{\dagger,1}$, Xingjue Liao$^{\dagger,1}$ and Feifei Qian$^{*,1}$

\thanks{$^{1}$ Department of Electrical and Computer Engineering, University of Southern California, Los Angeles, CA, USA.

$^{\dagger}$These authors contributed equally to this work.

$^*$Corresponding author: Feifei Qian. Email: {\tt\footnotesize feifeiqi@usc.edu}}
}
\maketitle

\begin{abstract}
Legged locomotion on inclined granular terrain depends sensitively on terrain properties and robot gait. Yet, the effect of gait timing on locomotor performance remains poorly understood. Here, we investigate how intra-cycle leg timing affects the upslope locomotion of a hexapedal robot on granular surfaces inclined from $0^\circ$ to $15^\circ$. We find that appropriately selected slow phase timing substantially improves locomotion relative to constant-angular-velocity actuation, while the optimal slow-phase center shifts toward smaller leg angles. In contrast to previous findings on level granular terrain, our single-leg force measurements show that this shift cannot be explained by the angular location of peak normal force. Instead, changing the slow-phase center reshapes the temporal combination of normal and shear forces generated by the phase-shifted tripod groups. Comparison of the reconstructed force profiles indicates that effective propulsion occurs when propulsive shear force coincides with sufficient terrain-generated normal support. A force-based model that combines granular resistive force theory with this propulsion criterion reproduces the observed dependence of locomotor performance and optimal gait timing on surface inclination, providing a framework for selecting optimal gait timing in challenging deformable environments.
\end{abstract}

\section{Introduction}\label{sec:Intro} 
Recent advances in mechanical design, sensing, control, and autonomy have substantially improved the mobility and versatility of legged robots~\cite{hutter2017anymal,bledt2018cheetah3,hwangbo2019learning,miki2022learning}. These advances have expanded their potential for applications including industrial inspection and search and rescue~\cite{hutter2017anymal,betta2024people}, as well as planetary exploration~\cite{arm2019spacebok,kolvenbach2022traversing}. Despite this progress, robust locomotion over deformable terrain remains challenging. Loose granular substrates can yield and rearrange under loading, leading to substantial leg penetration and slip~\cite{li2009sensitive,qian2013walking,qian2015principles}. These effects become more severe on granular slopes, where reduced shear resistance and increasing downslope loading can promote backward slip and eventually locomotion failure~\cite{kolvenbach2022traversing,liao2026failure}. On inclined granular terrain, increasing surface inclination alters the internal stress state of the substrate, reducing the available shear resistance and promoting downslope yielding and slip; insufficient terrain support can also lead to excessive sinkage and unstable leg--terrain contact~\cite{marvi2014sidewinding,kolvenbach2022traversing,liao2026failure}. These effects reduce locomotion robustness and can ultimately lead to slope-dependent locomotion failure~\cite{gosyne2018bipedial,liao2026failure}. 

Robophysics studies have demonstrated that understanding terrain–robot interactions can enable relatively simple strategies for improving locomotion on granular slopes~\cite{aguilar2016review}. For example, it was found that sidewinding robots can increase shear resistance forces and reduce slip on granular slopes by increasing the length of the body in contact with the substrate as inclination increases~\cite{marvi2014sidewinding}. Similarly, for bipedal locomotion, slope-dependent adjustment of step length and center-of-mass height has been shown to improve climbing success~\cite{gosyne2018bipedial}, while recent work has demonstrated that foot-driven terrain manipulation through cleated foot design can further improve locomotion on steep granular slopes~\cite{kerimoglu2026robust}. 
More recent work has further identified reduced shear resistance, delayed leg anchoring, and increased backward slip as important mechanisms underlying locomotion degradation and failure on granular slopes~\cite{liao2026failure}. 

While these studies demonstrate the importance of adapting robot–terrain interactions to surface inclination, they have largely focused on contact geometry, body posture, or foot design. The role of intra-cycle leg timing and coordination between multiple legs on granular slopes remains less well understood. On level granular terrain, it was found that locomotor performance is strongly influenced by limb kinematics and interaction speed~\cite{li2009sensitive,li2010effect}. In particular, Buehler-clock timing -- the placement of slow leg rotation phase within each gait cycle -- can strongly influence robot speed by changing how the slow phase overlaps with the force-generating portion of the leg trajectory~\cite{li2010effect}. Granular resistive force theory further provides a means of predicting the net force on complex intruders by integrating local resistive stresses over their surfaces~\cite{li2013terradynamics}. However, it remains unclear how the optimal Buehler-clock timing changes with surface inclination and whether the performance of an alternating-tripod gait can be explained from isolated single-leg forces or instead depends on the temporal combination of forces generated by the two phase-shifted tripod groups. 

In this study, we investigate how Buehler-clock timing governs the upslope locomotion of a hexapedal robot on granular surfaces. We systematically vary the center of the slow phase across four surface inclinations and show that the optimal timing shifts toward earlier leg angles as inclination increases. Single-leg force measurements demonstrate that this shift cannot be explained by a corresponding shift in the angle of peak normal force. We then reconstruct the combined normal and shear forces generated by the two phase-shifted tripod groups and show that gait timing controls when propulsive shear force coincides with sufficient normal support. By relating these force relationships to robot displacement on inclined granular media, we reveal the mechanism governing locomotion performance on granular slopes, and develop a predictive model that reproduces the dependence of locomotor performance and optimal gait timing on surface inclination and robot parameters, providing a basis for terrain-aware gait and robot-design optimization.
\section{Materials and Methods}
\label{sec:materials_methods}

\subsection{Robotic Platform and Gait}
\label{sec:robotic_platform}

A small RHex-type~\cite{saranli2001rhex} robot (Fig.~\ref{Fig.materials_methods}a) was used to investigate locomotion on level and inclined granular surfaces~\cite{saranli2001rhex}. The robot body measured 22~cm in length and 10~cm in width, with a total mass of 1.5~kg. It was equipped with six C-shaped rotary legs, each with a radius, $R$, of 3~cm, a width of 0.8~cm, and an arc span of $225^\circ$. The angular positions of the robot legs, $\phi$, were controlled using a microcontroller (Arduino Uno), and actuated by servo motors (Dynamixel XL330). The motor axles were mounted at a hip height, $h$, of 2~cm, measured vertically from the bottom of the body (Fig.~\ref{Fig.materials_methods}d). 

During all locomotion experiments, the robot used an alternating tripod gait~\cite{saranli2001rhex} in which the six legs were divided into two groups, as color-coded in Fig.~\ref{Fig.materials_methods}a. The three legs within each group moved synchronously, while the two groups moved with a phase offset of half a gait cycle. The alternating tripod gait was biologically inspired by insect locomotion~\cite{full1991mechanics,humeau2019locomotion} and has been widely used in studies of legged-robot locomotion on granular media~\cite{li2009sensitive, qian2015principles, liao2026failure}.

The periodic rotary motion of each leg was generated using a Buehler clock~\cite{saranli2001rhex}. Instead of rotating at a uniform angular velocity, each leg rotated slowly with angular speed $\omega_s$ through a prescribed angular interval and rapidly with angular speed $\omega_f$ through the remainder of the cycle, producing the piecewise-linear relationship between \(\phi\) and normalized time \(t/T\) shown in Fig.~\ref{Fig.materials_methods}c. The orange shaded slow interval (referred to as the slow phase hereafter) could be fully defined using the Buehler clock timing parameters, including its angular range $\phi_s$, duty factor $d_c$, and angular center $\phi_0$.

\subsection{Robot Locomotion Experiments}
\label{sec:locomotion_experiments}
\begin{figure}[!t]
\centering
\includegraphics[width=\columnwidth]{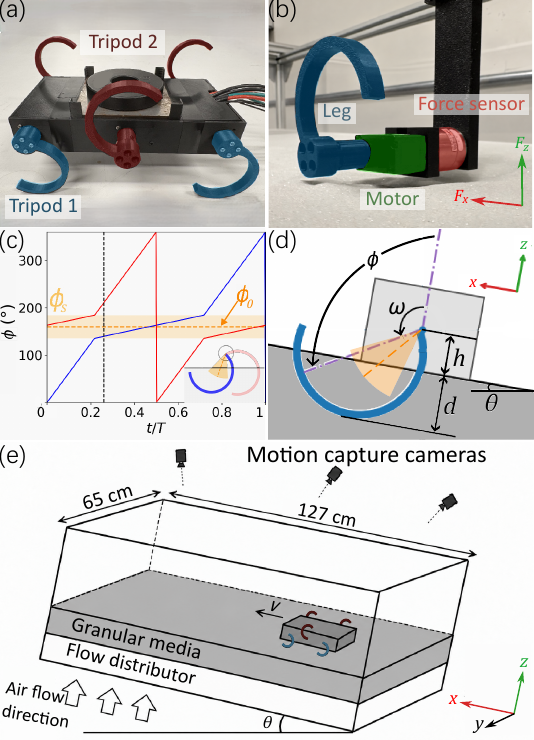}
\caption{Robot and experimental setup.
\textbf{(a)} RHex-type robot used in this study. The blue and red legs indicate the two alternating tripod groups. 
\textbf{(b)} Single-leg force-profile measurement setup. The force components $F_x$, $F_y$, and $F_z$ were measured along the fore-aft, lateral, and upward directions, respectively.
\textbf{(c)} Buehler-clock trajectories for the two tripod groups over one gait cycle, $t/T$. The blue and red curves show the leg angular positions of the two tripod groups, with a phase offset of $0.5T$ between them. The slow phase is defined by its angular range $\phi_s$, angular center $\phi_0$, and duty factor $d_c$, which specifies the fraction of the gait period occupied by the slow phase. The vertical black dashed line indicates the instant illustrated in the inset, which shows the corresponding angular configuration of the two tripod groups.
\textbf{(d)} A side-view diagram of the robot leg and kinematics. 
Here $h$ is the hip height,  
$d$ is the leg penetration depth, 
$\phi$ is the leg angular position, measured counterclockwise from the $+z$ direction to the line connecting the leg rotation axis to the center of curvature of the C-shaped leg; 
$\omega$ is the leg angular velocity, and $\theta$ is the surface inclination.
\textbf{(e)} Tiltable fluidized-bed setup used for robot locomotion experiments. The robot traveled in the upslope direction with velocity $v$, while three motion-capture cameras recorded its position in the coordinate frame $(x,y,z)$.}
\label{Fig.materials_methods}
\end{figure}

Robot locomotion experiments were conducted in a 127~cm $\times$ 65~cm tiltable fluidized bed~\cite{qian2013automated, jin2019preparation}, as illustrated in Fig.~\ref{Fig.materials_methods}e. The fluidized bed was filled with \(300~\mathrm{\text\textmu m}\)-diameter glass beads (Grainger) to a depth of 10~cm. This depth was selected to minimize boundary effects~\cite{stone2004bottom} on the interaction between the robot legs and the granular substrate. Glass beads are widely used as a model granular substrate~\cite{qian2013walking,schiebel2020mitigating,sharpe2015controlled,nayak2025burrowing,liao2026failure}, as they produce qualitatively similar rheology with sand, while their uniform size and simple geometry facilitate the preparation of repeatable packing states~\cite{mueth1998force}.

Three motion-capture cameras (OptiTrack Prime 13W) were installed around the fluidized bed to record the three-dimensional position (Fig. \ref{Fig.materials_methods}e, $x$-fore-aft, $y$-lateral, $z$-surface normal) of the robot at 240 frames per second. An additional video camera was positioned beside the sand bed to record side-view video of each trial at 30 frames per second.

Before each trial, air was passed upward through the base of the sand bed to fluidize the glass beads and reset the compaction and level of the substrate. The airflow was then gradually ramped down to zero, allowing the particles to settle into a flat, loosely packed state~\cite{li2009sensitive,qian2013automated}. The sand bed was then tilted to the prescribed inclination, $\theta$ (Fig. \ref{Fig.materials_methods}e). Subsequently, the robot was placed at a fixed initial position with its longitudinal axis aligned with the upslope direction (Fig. \ref{Fig.materials_methods}e, $x$-axis).



Using this data collection protocol, we performed two sets of experiments to evaluate the effect of Buehler-clock timing on robot locomotion performance: (1) Buehler-clock actuation with a systematically varied slow-phase center, to determine how slow-phase timing affects locomotion and identify the optimal timing at each inclination, and (2) constant-angular-velocity actuation as a baseline to evaluate the benefit of nonuniform leg rotation speed. Both sets of experiments were conducted at surface inclinations of $\theta=0^\circ$, $5^\circ$, $10^\circ$, and $15^\circ$.

For the Buehler-clock experiments, the slow-phase center was initially varied from $\phi_0=140^\circ$ to $190^\circ$ in increments of $10^\circ$. Additional values were tested near the high-displacement region to refine the resolution: $\phi_0=165^\circ$ and $175^\circ$ at $\theta=0^\circ$, $165^\circ$ at $\theta=5^\circ$, $155^\circ$ and $165^\circ$ at $\theta=10^\circ$, and $145^\circ$, $155^\circ$, and $165^\circ$ at $\theta=15^\circ$. Each tested condition was repeated three times. For all trials, the gait period was fixed at $T=\pi~\mathrm{s}$, with a phase offset of $0.5T$ between the two alternating leg groups (Fig.~\ref{Fig.materials_methods}c, blue and red). The slow-phase angular range was fixed at $\phi_s=48.7^\circ$, and the slow phase occupied half of the gait period, corresponding to a duty factor of $d_c=0.5$.

For the constant-angular-velocity experiments, each leg rotated at $\omega=114.6^\circ,\mathrm{s^{-1}}$ ($2~\mathrm{rad,s^{-1}}$). This angular velocity gives the same gait period, $T=\pi~\mathrm{s}$, and the same average angular velocity as the Buehler-clock gait, allowing the effect of intra-cycle velocity modulation to be evaluated relative to a constant-velocity baseline.

For all trials, the robot completed 10 gait cycles, and the displacement between its positions in the initial and final motion-capture frames was used to quantify locomotion performance and compare the effects of surface inclination, slow-phase center, and actuation strategy. 

\subsection{Single-Leg Force-Profile Measurements}
\label{sec:force_profile_measurements}

To understand how the robot legs interacted with the granular substrate during locomotion, we conducted single-leg force-profile measurements. A C-shaped leg driven by a servo motor was mounted at a fixed height above the granular surface, whereas a six-axis force--torque sensor (ATI, Nano25) measured the granular resistive force exerted on the leg (Fig.~\ref{Fig.materials_methods}b). Measurements were conducted at inclination angles of $\theta=0^\circ$, $5^\circ$, $10^\circ$, and $15^\circ$, corresponding to the same inclinations used in the locomotion experiments. Before each measurement, the granular bed was prepared using the same fluidization and settling procedure as in the robot locomotion experiments. After the sand bed reached the prescribed inclination, the force sensor was biased to zero before the leg began rotating, thereby removing the static gravitational contribution of the leg and mounting assembly from the measured forces.

Two sets of force-profile measurements were conducted. In the first set, the leg was driven directly using the complete Buehler-clock trajectory employed in the robot locomotion experiments. In the second set, the leg was rotated at the constant angular velocities $\omega_s$ and $\omega_f$, corresponding to the slow and fast phases of the Buehler-clock gait, respectively. During each trial, the synchronized force components (Fig.~\ref{Fig.materials_methods}b, $F_x$, and $F_z$) and leg angular position $\phi$ were recorded.

\section{Results and discussion}\label{sec:results} 
\subsection{Effect of Slow-Phase Timing on Robot Locomotion Performance}
\label{sec:locomotion_results}

\begin{figure}[t]
\centering

\includegraphics[width=\columnwidth]{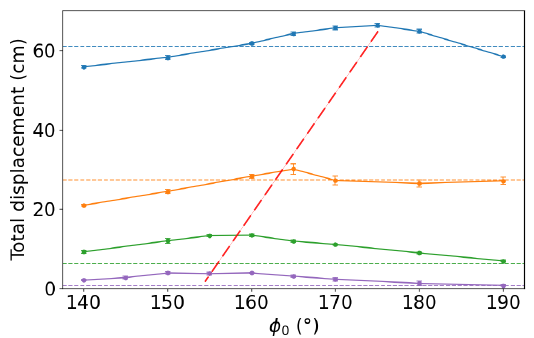}

\caption{
Robot locomotion performance under Buehler-clock and
constant-angular-velocity actuation on inclined granular surfaces.
Total displacement over ten gait cycles is shown as a function of
the slow-phase center, $\phi_0$, for surface inclinations of
$0^\circ$, $5^\circ$, $10^\circ$, and $15^\circ$
(blue, orange, green, and purple, respectively).
Solid curves show the mean across three trials with
standard-deviation error bars, while dashed horizontal lines indicate
the corresponding constant-angular-velocity baselines.
The red dashed line highlights the shift of the optimal slow-phase center
toward smaller $\phi_0$ as surface inclination increases.
}
\label{fig:locomotion_results}
\end{figure}

Across all tested inclinations, robot center-of-mass (CoM) displacement varied non-monotonically with the slow phase center, $\phi_0$, exhibiting a localized high-displacement region rather than a monotonic increase or decrease (Fig. \ref{fig:locomotion_results}). As the surface inclination, $\theta$, increased, the best-performing slow-phase center, $\phi_0^*$, shifted systematically toward smaller values of $\phi_0$. On the level granular surface, $\phi_0^*=175^\circ$, corresponding to a slow phase centered when the C-leg's center of curvature was nearly directly below the hip (Fig. \ref{fig:locomotion_results}, blue). As the surface inclination increased, $\phi_0^*$ shifted to $165^\circ$ at $\theta=5^\circ$ (Fig. \ref{fig:locomotion_results}, orange), meaning that the slow phase began earlier before the leg reached this downward orientation, and further shifted to $160^\circ$ and between $150 - 160^\circ$ at $\theta=10^\circ$ and $15^\circ$, respectively (Fig. \ref{fig:locomotion_results}, green and purple). 

\begin{figure}
    \centering
\refstepcounter{table}
\label{tab:locomotion_results}
{\footnotesize
\textbf{TABLE~\Roman{table}}\\[-1mm]
\textsc{Locomotion Performance and Optimal Gait Timing}
}
\resizebox{\columnwidth}{!}{%
\begin{tabular}{c c c c c}
\hline
$\theta$ ($^\circ$) &
$\phi_0^*$ ($^\circ$) &
$D_{\mathrm{bclock}}$ (cm) &
$D_{\mathrm{constant}}$ (cm) &
Gain (\%) \\
\hline
0  & 175      & $66.41 \pm 0.43$ & $61.10 \pm 0.46$ & 8.7   \\
5  & 165      & $30.07 \pm 1.37$ & $27.27 \pm 0.61$ & 10.3  \\
10 & 150--160 & $13.40 \pm 0.34$ & $6.34 \pm 0.29$  & 111.4 \\
15 & 150--160 & $3.85 \pm 0.08$  & $0.71 \pm 0.23$  & 442.3 \\
\hline
\end{tabular}%
}
\end{figure}
Although the absolute CoM displacement decreased rapidly as the surface inclination increased, the relative benefit of an appropriately selected Buehler-clock gait increased with inclination. Relative to constant-angular-velocity actuation, the best-performing Buehler-clock gait increased displacement by approximately 8.7\% at $0^\circ$, 10.3\% at $5^\circ$, 111.4\% at $10^\circ$, and 442.3\% at $15^\circ$ (Table~\ref{tab:locomotion_results}). The constant-angular-velocity gait became ineffective on the steepest surface (with a displacement of only 0.71 cm on 15$\degree$), whereas the Buehler-clock gait retained measurable upslope locomotion (3.85 cm displacement on 15$\degree$). These results indicate that controlling the intra-cycle distribution of leg angular velocity becomes increasingly important as the granular terrain becomes more challenging. We next investigate the physical mechanism underlying the observed shift in optimal slow phase timing. 

\subsection{Single-Leg Normal-Force Peak Position Does Not Explain the Shift in Optimal Slow-Phase Center}
\label{sec:single_leg_force_peak}

Prior work on level granular terrain has suggested that locomotor performance was maximized when the slow phase overlapped with the angular interval in which the granular penetration resistive force from the sand was the largest~\cite{li2010effect}. This overlap promotes a solid-like response of the granular substrate: when the penetration resistance becomes sufficient to support and accelerate the robot, further leg penetration is no longer needed and the leg can instead rotate against the solidified region to lift and advance the body. Because slower leg rotation also reduces the acceleration required from the substrate, positioning the slow phase within the high-resistance portion of the leg cycle increases the angular interval over which this supported rotary-walking motion can occur, thereby increasing locomotion speed.
Motivated by these findings, we first examined whether the shift in optimal timing observed on inclined surfaces could be explained by an inclination-dependent shift in the single-leg force profile. We therefore proposed the following hypothesis: 

\noindent\textbf{Hypothesis 1.} Increasing the surface inclination shifts the angular position corresponding to the peak normal resistive force exerted by the granular substrate, resulting in the observed shift in the location of the best-performing slow-phase center, $\phi_0^*$. 


If this hypothesis is correct, the observed shift of $\phi_0^*$ toward smaller leg angles at higher inclinations would keep the slow phase aligned with the range of leg angles over which large normal forces are generated. 

\begin{figure}[htbp]
    \centering
    \includegraphics[width=\linewidth]{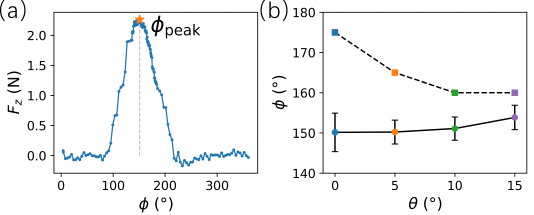}
    \caption{
    Single-leg peak normal-force location and optimal slow-phase center as functions of surface inclination.
    \textbf{(a)} Representative single-leg normal-force profile as a function of leg angle. The orange star marks the maximum $F_z$, and the corresponding leg angle is defined as the peak-force angle, $\phi_{\mathrm{peak}}$; the light vertical dashed line indicates its angular location.
    \textbf{(b)} Comparison between the mean peak-force angle, $\phi_{\mathrm{peak}}$ (solid black line with circular markers), and the optimal slow-phase center, $\phi_0^*$ (black dashed line with square markers), as functions of surface inclination. Error bars for $\phi_{\mathrm{peak}}$ indicate the standard deviation across the tested slow-phase centers. Marker colors correspond to $\theta=0^\circ$, $5^\circ$, $10^\circ$, and $15^\circ$ in blue, orange, green, and purple, respectively.
    }
    \label{fig:single_leg_peak_angle}
\end{figure}

To test Hypothesis 1, we measured the angular position of the peak single-leg normal force at each inclination ($0^\circ$, $5^\circ$, $10^\circ$, and $15^\circ$), and compared it with the best-performing slow-phase center measured from the locomotion experiments. Figure~\ref{fig:single_leg_peak_angle} shows the peak-force angles obtained from the individual force profiles, together with their mean and sample standard deviation at each surface inclination. For each force profile, the peak-force angle, $\phi_{\mathrm{peak}}$, was defined as the leg angle at which $F_z$ reached its maximum value (Fig.~\ref{fig:single_leg_peak_angle}a, orange star).

Our measurements show that the mean peak-force angle, $\phi_{\mathrm{peak}}$, remained approximately constant as the surface inclination increased from 0$\degree$ to 15$\degree$ (Fig.~\ref{fig:single_leg_peak_angle}b, circular markers). In contrast, the best-performing slow-phase center, $\phi_0^*$, shifted significantly toward smaller leg angles with increasing inclination (Fig.~\ref{fig:single_leg_peak_angle}b, square markers). These results do not support Hypothesis 1: the inclination-dependent shift in optimal slow-phase timing cannot be explained by a corresponding shift in the peak of the single-leg normal-force profile. The optimal timing of the Buehler clock, therefore, is not determined solely by aligning the slow phase with a single-leg force maximum. 

\subsection{Slow-Phase Timing Reshapes the Combined Force Profiles of Phase-Shifted Legs}
\label{sec:overlapped_force_profiles}

Inspection of the gait kinematics showed that both alternating tripod groups were simultaneously engaged with the granular substrate during part of each gait cycle (Fig.~\ref{Fig.materials_methods}c, inset). Previous studies of two simultaneously intruding objects have shown that the granular resistance generated by neighboring intruders can depend strongly on their separation, reflecting interactions mediated through the surrounding granular material~\cite{umbanhowar2009enhanced,pravin2021effect}. During this interval, the net force acting on the robot therefore depends on the temporal combination of the forces generated by the two phase-shifted tripod groups. This motivated us to consider whether the temporal combination of forces generated by the phase-shifted tripod groups, rather than the force from isolated legs, could explain the shift in optimal timing. We therefore formulated the following hypothesis:

\noindent\textbf{Hypothesis 2.} The slow-phase center controls the temporal alignment of the normal and shear forces generated by the two phase-shifted tripod groups. This combined force profile determines the optimal slow-phase timing by governing when propulsive shear force coincides with sufficient normal support.

If this hypothesis is correct, the value of $\phi_0$ that maximizes the propulsive effect of the combined shear force during periods of sufficient normal support should correspond to the experimentally measured $\phi_0^*$.%

\begin{figure}[t]
    \centering
    \includegraphics[width=\columnwidth]{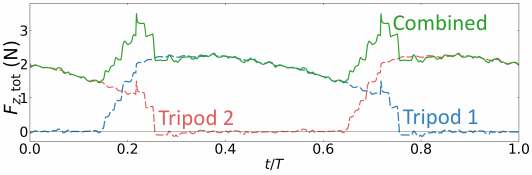}
    \caption{Reconstruction of the combined normal-force profile,
    $F_{z,\mathrm{tot}}$, over one gait cycle on a $10^\circ$ granular slope
    at $\phi_0=160^\circ$. The blue and red dashed curves show the individual
    normal-force contributions from the two legs separated by $0.5T$, and the
    green solid curve shows their sum.}
    \label{fig:combined_force}
\end{figure}

To evaluate Hypothesis 2, we reconstructed the combined force generated by a representative pair of phase-shifted legs, with one leg from each alternating tripod group, using the single-leg force profiles measured at the constant slow- and fast-phase angular velocities (Fig.~\ref{fig:combined_force}, Tripod 1 and Tripod 2). The two legs followed identical Buehler-clock trajectories separated by a phase offset of $0.5T$. At each instant, the force generated by each leg was obtained from either the slow-velocity or fast-velocity force profile, according to its current phase, and evaluated at its instantaneous leg angle. 
The forces from the base leg and the phase-shifted leg were summed to obtain the combined force contribution (Figure~\ref{fig:combined_force}, Combined):



\begin{equation}
F_{i,\mathrm{tot}}(t;\phi_0)
=
F_i(t;\phi_0)
+
F_i\left(t+\frac{T}{2};\phi_0\right),
\qquad i\in\{x,z\}.
\label{eq:combined_force}
\end{equation}

 Here, $F_x(t;\phi_0)$ and $F_z(t;\phi_0)$ are the reconstructed shear and normal forces generated by the base leg, respectively, and the second terms are the corresponding forces generated by the phase-shifted leg\footnote{For clarity, the dependence of these forces on surface inclination \(\theta\) is omitted from the notation.}. 
 
\begin{figure}[htbp]
\centering
\includegraphics[width=\columnwidth]
{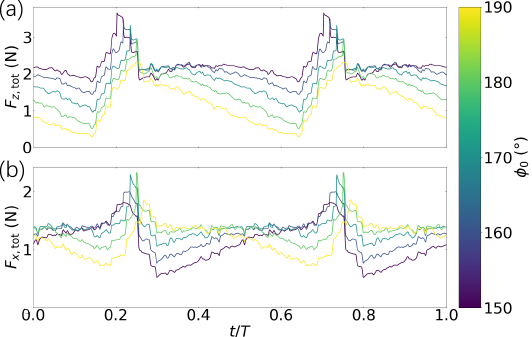}
\caption{
Reconstructed combined normal- and shear-force profiles on a $10^\circ$ granular slope for five slow-phase centers, $\phi_0$. For each $\phi_0$, the forces generated by two legs following identical Buehler-clock trajectories with a phase offset of $0.5T$ were reconstructed from the measured single-leg constant-angular-velocity force profiles corresponding to the slow- and fast-phase velocities and then summed. \textbf{(a)} Combined normal force, $F_{z,\mathrm{tot}}$. \textbf{(b)} Combined shear force, $F_{x,\mathrm{tot}}$. Curve color denotes the slow-phase center, with $\phi_0$ increasing from $150^\circ$ to $190^\circ$ as indicated by the color bar.
}
\label{fig:combined_force_withdiffspc}
\end{figure}

Fig.~\ref{fig:combined_force_withdiffspc} compares the reconstructed combined force profiles for different values of $\phi_0$. Among the tested values, $\phi_0=150^\circ$ produced the largest peak in $F_{z,\mathrm{tot}}$ (Fig.~\ref{fig:combined_force_withdiffspc}, purple), whereas the peak magnitude progressively decreased as $\phi_0$ increased to $190^\circ$ (Fig.~\ref{fig:combined_force_withdiffspc}, yellow). In contrast, the corresponding changes in $F_{x,\mathrm{tot}}$ varied non-monotonically with $\phi_0$, indicating that the combined normal and shear forces respond differently to changes in slow-phase timing.

The effect of $\phi_0$ on the reconstructed force profiles arises from both the velocity-dependent leg–terrain interaction and the angle--time mapping imposed by the Buehler clock. Changing $\phi_0$ determines which angular interval of the leg trajectory is traversed slowly and which measured velocity-dependent force profile is applied at each leg angle. It also determines how long the leg remains within each angular interval. 
For example, when the slow phase overlaps with the angular region of high force in the measured single-leg force–angle profile, the leg remains near its peak force for a longer portion of the gait cycle (Fig.~\ref{fig:combined_force}, Tripod 1). When the contributions from the two phase-shifted legs are summed, these $\phi_0$-dependent changes in the magnitude and temporal distribution of the single-leg forces alter how the two force histories combine, thereby changing the magnitude, width, and shape of the total force profile.

These results suggest that the locomotor effect of the combined force depends not only on the magnitude of the shear force but also on whether favorable shear force coincides with sufficient normal support. We therefore propose that this coincidence defines an effective propulsion interval during which the robot accelerates upslope. In the following section, we test this interpretation by relating the combined normal and shear force profiles to locomotor performance.

\subsection{Predicting Step Length from the Effective Propulsion Window}
\label{sec:PredictingStepLength} 

In this section, we relate the phase-modulated total forces to the slope- and $\phi_0$-dependent locomotor performance
observed in Fig.~\ref{fig:locomotion_results}. We evaluate two assumptions: (1) effective propulsion occurs
only when the combined normal force provides sufficient support, and the duration of this support sets the propulsion window; and (2) the combined shear force within this window determines the forward displacement. 
We first identify the propulsion window from the measured robot kinematics and compare its boundaries with the reconstructed combined normal force. We then integrate the reconstructed shear force within this experimentally identified window to estimate robot displacement and compare the estimate with the locomotion measurements. 

\subsubsection{Determining propulsion window from combined normal force}



\begin{figure}[htbp]
\centering
\includegraphics[width=1\linewidth]
{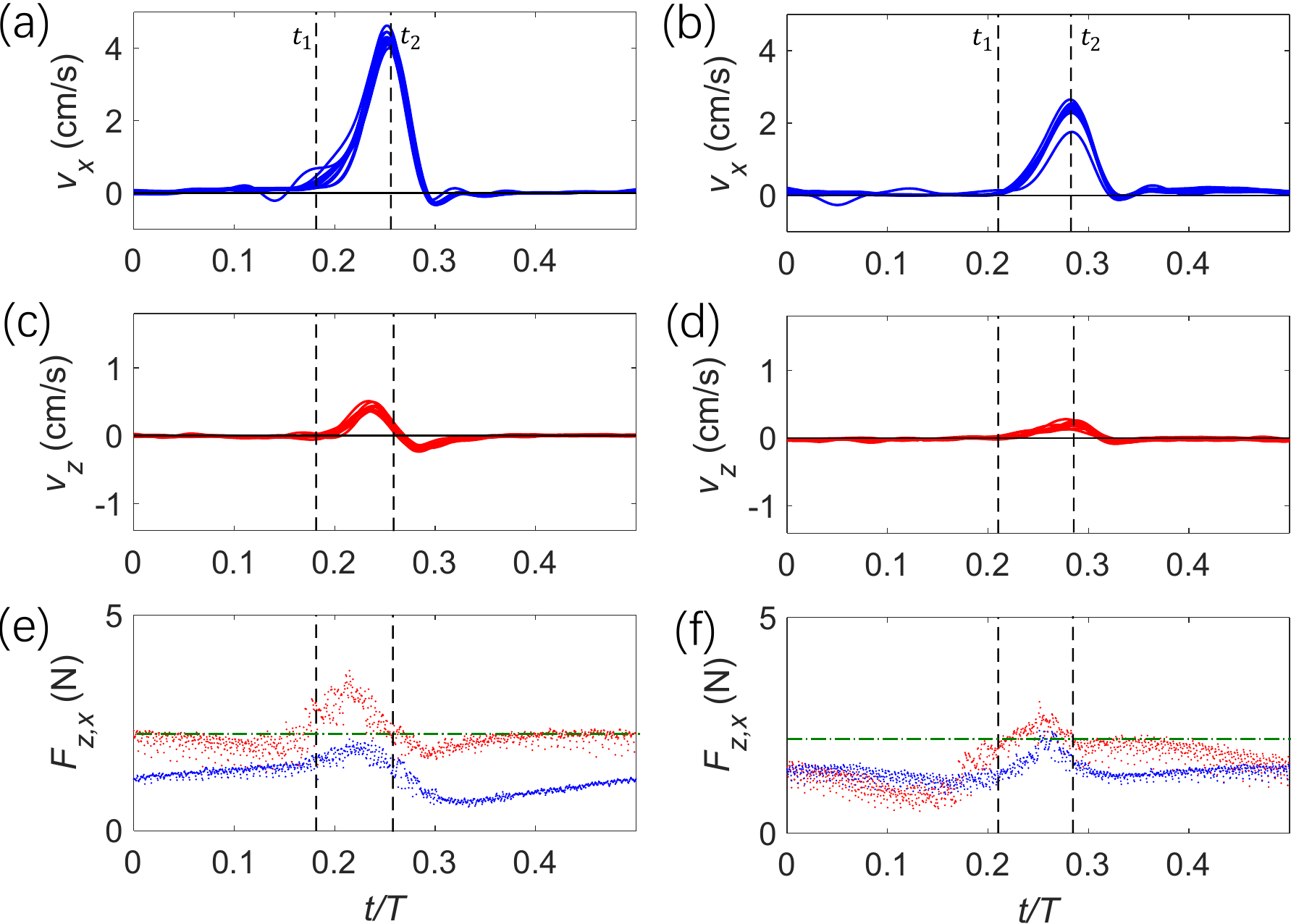}
\caption{Robot speed, reconstructed granular resistive forces, and corresponding effective propulsion window, for locomotion on a $10^\circ$ granular slope. Panels \textbf{(a,c,e)} show trials with $\phi_0=155^\circ$, and panels \textbf{(b,d,f)} show trials with $\phi_0=180^\circ$. 
\textbf{(a, b)} Robot CoM upslope velocity $v_x$, \textbf{(c, d)} Robot CoM surface-normal velocity, $v_z$ and \textbf{(e, f)} Reconstructed combined normal force, where the red markers denote the normal force $F_z$ and the blue markers denote shear force $F_x$. The green horizontal line represents the predicted normal-support threshold,
$F_{\mathrm{th}}(10^\circ)$.
The vertical dashed lines mark $t_1$ and $t_2$, the onset of upslope acceleration and deceleration, respectively. The interval [$t_1$, $t_2$] defines the effective propulsion window.} 
\label{fig:Speed}
\end{figure}

To determine the propulsion window, we first examine how terrain-generated normal support relates to the timing of effective propulsion within each step (half stride). From the experimentally measured robot motion, we identify $t_1$ and $t_2$ as the onset of upslope acceleration and deceleration, respectively (Fig.~\ref{fig:Speed}a, b). The corresponding surface-normal velocity in Fig.~\ref{fig:Speed}c, d shows that this interval also coincides with the primary vertical motion of the body, suggesting that forward acceleration occurs while the legs provide sufficient support to lift and propel the robot. 

We define the experimentally identified effective propulsion window of each step as
\begin{equation}
    \mathcal{T}_{\mathrm{exp}}(\phi_0,\theta)=[t_1,t_2].
\end{equation}
Comparing this kinematically identified window with the reconstructed force profiles reveals a corresponding peak in the combined normal force. As shown in Fig.~\ref{fig:Speed}e, f, $t_1$ occurs as $F_{z,\mathrm{tot}}$ (Fig.~\ref{fig:Speed}e, f, red) reaches a characteristic level, $F_{\mathrm{th}}$ (Fig.~\ref{fig:Speed}e, f, green dashed line), whereas $t_2$ occurs as it falls below a similar level. This correspondence suggests that effective propulsion begins only after sufficient terrain-generated normal support is established and ends when this support is lost.

This interpretation is consistent with previous studies of C-legged locomotion on granular media. During rotary walking, the legs must generate sufficient penetration resistance force to support the robot body, reducing body--ground contact and allowing leg rotation against the substrate to advance the body~\cite{li2009sensitive,li2013terradynamics}.
As such, the effective propulsion window, $\mathcal{T}$, can be identified from the normal force support:   
\begin{equation}
    \mathcal{T}(\phi_0,\theta)
    =
    \left\{
    t:F_{z,\mathrm{tot}}(t;\phi_0,\theta)>F_{\mathrm{th}}(\theta)
    \right\},
\label{eq:window}
\end{equation}
In this section, we use the kinematically identified window $\mathcal{T}_{\mathrm{exp}}$ to evaluate the relationship between shear force and displacement.  A quantitative method for estimating $F_{\mathrm{th}}$ and thereby predicting $\mathcal{T}$ is presented in Sec.~\ref{sec:predictingThreshold}.



\subsubsection{Estimating robot displacement from integrated shear force during propulsion}

The normal-force criterion determines when effective propulsion can occur, but the duration of this window alone does not determine how far the robot advances. Within this window, the shear component of the leg--terrain interaction acts along the direction of travel and contributes to the robot's upslope acceleration. The resulting robot displacement should therefore depend on both the magnitude and duration of the shear force. To evaluate this relationship, we integrate the reconstructed shear force over the experimentally identified propulsion window, $\mathcal{T}_{\mathrm{exp}}=[t_1,t_2]$, and estimate the net upslope acceleration as
\begin{equation}
a_x(t;\phi_0,\theta)
= \frac{n\eta}{M}F_{x,\mathrm{tot}}(t;\phi_0,\theta)-g\sin\theta ,
\end{equation}

\noindent and the resulting displacement per step (half stride) as 
\begin{equation}
s_p(\phi_0,\theta)
=
\int_{t_1}^{t_2}
\int_{t_1}^{t}
a_x(\tau;\phi_0,\theta)\,d\tau\,dt. 
\label{eq:step_prediction}
\end{equation}
Here, $n=3$ is the number of legs within one alternating tripod, $M$ is the robot mass, $g\sin\theta$ is the downslope gravitational acceleration, $\eta=0.5$ is the effective propulsion coefficient, representing the fraction of the reconstructed leg-generated shear force that contributes to robot acceleration, identified following~\cite{maladen2011undulatory, schiebel2020mitigating}. 
The robot forward speed was assumed to be 0 prior to the propulsion window.

\begin{figure}[htbp]
\centering
\includegraphics[width=0.9\linewidth]
{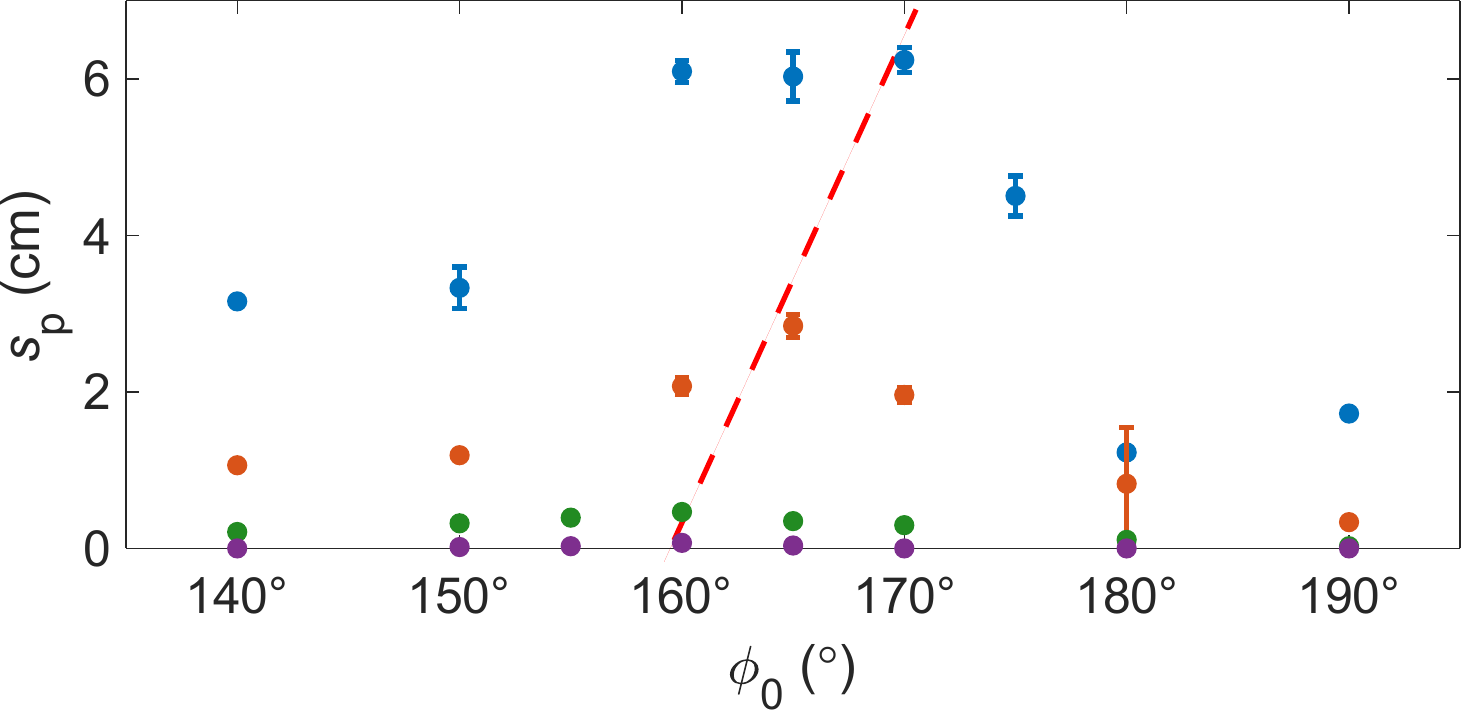}
\caption{
Predicted robot displacement from the reconstructed shear force during the effective propulsion window. The blue, orange, green, and purple markers correspond to $\theta=0^\circ$, $5^\circ$,
$10^\circ$, and $15^\circ$, respectively. The red dashed line highlights the shift of the highest-displacement prediction toward smaller $\phi_0$ as surface inclination increases, consistent with the experimentally observed shift in optimal slow-phase center shown in Fig.~\ref{fig:locomotion_results}. 
} 
\label{fig:Window}
\end{figure}

The estimated robot displacement, $s_p$, is shown in Fig.~\ref{fig:Window} as a function of the slow-phase center $\phi_0$. We noticed two trends. First, $s_p$ decreases substantially as $\theta$ increases, consistent with the overall degradation of locomotor performance observed experimentally in Fig.~\ref{fig:locomotion_results}. Second, at each inclination, $s_p$ depends non-monotonically on $\phi_0$ and exhibits a preferred range of slow-phase centers. This high-step-length region shifts toward smaller $\phi_0$ as the surface inclination increases (Fig.~\ref{fig:Window}, red dashed line), qualitatively reproducing the experimentally observed shift in the optimal slow-phase center (Fig.~\ref{fig:locomotion_results}).


Together, these results are consistent with the proposed mechanism: sufficient combined normal force sets the effective propulsion window, while the combined shear force acting within this window determines the resulting forward displacement. In the next section, we use this mechanism to predict the optimal gait timing. 
\section{Model generalization and broader applicability}
\label{sec:General-Shape}
While the analysis in Sec.~\ref{sec:PredictingStepLength} explains the observed locomotion performance, its reliance on measured force profiles and robot kinematics limits its application to conditions that have already been tested. To achieve a more general framework that predicts locomotion performance directly from the robot, gait, and terrain parameters, we use granular resistive force theory (RFT) to predict the leg-terrain force profiles, and then use the predicted forces to determine the effective propulsion window and estimate the resulting step length.

\subsection{Estimating the force profile}
\label{sec:RFT}
Granular RFT allows prediction of the resistive force acting on an intruder by integrating local granular stresses over its interacting surface~\cite{li2013terradynamics}. For the C-shaped leg, the normal and shear components are written as \begin{equation}
    F^{R}_{z,x}
    =
    \int_S
    \alpha_{z,x}(\beta,\gamma)|z|\,dA,
    \label{eq:RFT}
\end{equation}
where $S$ is the portion of the leg interacting with the granular substrate, $z$ is the local penetration depth, and $\alpha_{z,x}(\beta,\gamma)$ are the RFT stress coefficients associated with each individual leg element, with values obtained from the generic stress relations reported in~\cite{li2013terradynamics}. The angles $\beta$ and $\gamma$ describe the local orientation and direction of motion of each leg element, respectively. Because the leg trajectory is prescribed by the Buehler clock, the instantaneous position, orientation and velocity direction of each leg element can be determined directly from the gait parameters, allowing the force profile to be evaluated throughout a gait cycle.

The original granular RFT~\cite{li2013terradynamics} applied to level substrates. To estimate granular forces on inclined surfaces, we introduce dimensionless correction factors based on the penetration and shear resistance measurements reported in~\cite{liao2026failure}:
\begin{align}
    \alpha_x^*(\beta,\gamma,\theta)
    &= G_x(\theta)\alpha_x(\beta,\gamma), \\
    \alpha_z^*(\beta,\gamma,\theta)
    &= G_z(\theta)\alpha_z(\beta,\gamma).
    \label{eq:RFT_slope}
\end{align}
Here $\alpha_x$ and $\alpha_z$ are the RFT stress coefficients for level granular terrain, whereas $\alpha_x^*$ and $\alpha_z^*$ represent the slope-dependent stress coefficients. The dimensionless correction factor $G_x(\theta)$ and $G_z(\theta)$ represent the experimentally derived reduction in shear and normal penetration resistance, respectively
\footnote{Based on the terrain-resistance measurements in~\cite{liao2026failure}, we approximate the normal penetration resistance as independent of surface inclination over the tested range, such that $G_z(\theta) = 1$, while $G_x(\theta)$ decreases linearly from $G_x(0^\circ)=1$ to $G_x(15^\circ)=0.7$.}.

Once the resistive forces are computed for each leg, the alternating-tripod gait is incorporated using the same superposition procedure introduced in Sec.~\ref{sec:overlapped_force_profiles}. The RFT-predicted forces generated by a representative pair of legs separated by $T/2$ are summed to obtain the combined normal and shear forces, $F^{R}_{z,\mathrm{tot}}$ and
$F^{R}_{x,\mathrm{tot}}$, respectively.

\subsection{Estimating the Normal Support Threshold}
\label{sec:predictingThreshold}
With the force profiles predicted from RFT, the remaining step is to determine the normal-support threshold, $F_{\mathrm{th}}$, that defines the effective propulsion window. We calibrate this threshold without using measured robot kinematics by matching the force-based step-length estimate, \(s_p\), to the step length predicted by an established rotary-walking model for level granular terrain~\cite{li2009sensitive}.  

For level granular terrain, the step length of a C-legged robot during rotary walking can be approximated as
\begin{equation}
    s^* = 2\sqrt{R^2-(d+h-R)^2},
    \label{eq:level_step}
\end{equation}
where $R$ is the leg radius, $h$ is the hip height, and $d$ is the maximum leg penetration depth, which can be calculated from the robot and terrain parameters following~\cite{li2009sensitive}. 
We determine the level-ground threshold, $F_{\mathrm{th,0}}$, from a reference gait condition with $\theta=0^\circ$ and $\phi_0=160^\circ$. For each candidate value of $F_{\mathrm{th},0}$, the intersections of the threshold with the dominant peak of the RFT-predicted normal force profile, $F^R_{z,\mathrm{tot}}$, uniquely define the propulsion window $[t_1,t_2]$ through Eq.~\ref{eq:window}. The RFT-predicted shear force, $F^R_{x,\mathrm{tot}}$, is then integrated over this window using Eq.~\ref{eq:step_prediction} to obtain the step length, $s_p$. The threshold, $F_{\mathrm{th},0}$, is selected such that $s_p (F_{\mathrm{th},0}) = s^*$. As a simplifying assumption, we treat the effective support threshold as a property of the robot and terrain condition and therefore we assume that it is independent of $\phi_0$.

For inclined terrain, we approximate the change in the support requirement using the surface-normal component of gravity,
\begin{equation}
    F_{\mathrm{th}}(\theta)
    =
    F_{\mathrm{th},0}\cos\theta.
\end{equation}
The factor $\cos\theta$ accounts for the reduction in the surface-normal component of gravity as inclination increases. Substituting $F_{\mathrm{th}}(\theta)$ into the propulsion criterion in Eq.~\ref{eq:window} determines the predicted propulsion window for given $\phi_0$ and $\theta$. The corresponding RFT-predicted shear force within this window is then used in Eq.~\ref{eq:step_prediction} to estimate the corresponding step length.

\subsection{Generalization to Robot Design and Gait Optimization}
\label{sec:optimization}
The model developed above enables the estimation of locomotion performance for specific robot, gait, and terrain parameters without directly measuring robot motion under every condition. More generally, for a given robot and terrain condition, the model can
evaluate candidate Buehler-clock parameters and identify the slow-phase center that
maximizes the predicted step length,
\begin{equation}
    \phi_0^*
    =
    \underset{\phi_0}{\arg\max}\,
    s_p(\phi_0,\theta).
\end{equation}

\begin{figure}[htbp]
\centering
\includegraphics[width=0.9\linewidth]
{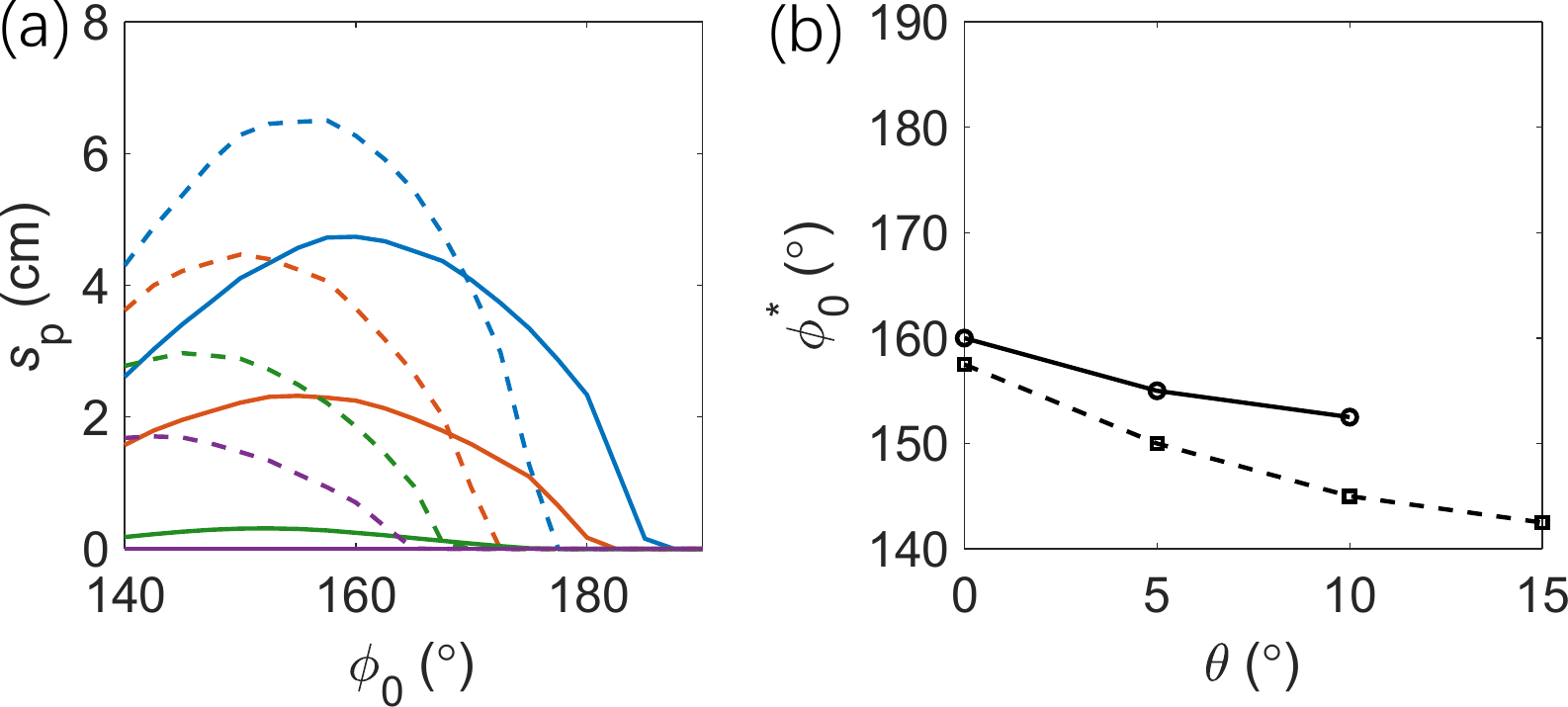}
\caption{
Predicted effect of leg radius on locomotion performance and optimal Buehler-clock timing.
\textbf{(a)} Predicted step length, $s_p$, as a function of the  slow-phase center $\phi_0$ for leg radii $R=3$~cm (solid lines) and  $R=4$~cm (dashed lines). The blue, orange, green, and purple curves 
correspond to $\theta=0^\circ$, $5^\circ$, $10^\circ$, and $15^\circ$, 
respectively.
\textbf{(b)} Optimal slow-phase center, $\phi_0^*$, as a function of  surface inclination for $R=3$~cm (solid line) and $R=4$~cm (dashed line). For $R=3$~cm at $\theta=15^\circ$, the model predicts zero displacement  for all tested values of $\phi_0$, so no optimal $\phi_0^*$ is reported.
}
\label{fig:FullPrediction}
\end{figure}

Here, we use the model to examine how leg radius affects predicted locomotion performance and optimal slow-phase timing. Fig.~\ref{fig:FullPrediction}a compares the predicted step length, $s_p$, for robots with leg radii $R=3$~cm and $R=4$~cm across different surface inclinations. The predicted optimal slow-phase centers, $\phi_0^*$, decrease with increasing surface inclination for both leg radii (Fig.~\ref{fig:FullPrediction}b). The model also predicts a smaller $\phi_0^*$ for the robot with larger legs, suggesting that optimal gait timing depends jointly on robot morphology and terrain inclination.

Thus, the framework provides a quantitative tool for relating terrain
mechanics and robot parameters to locomotion performance, and selecting terrain- and robot-dependent gait timing.

\section{CONCLUSION}
In this work, we investigated how Buehler-clock timing governs the locomotion of a RHex-type robot on inclined granular terrain. We found that the optimal slow-phase center shifts toward smaller leg angles as surface inclination increases. Single leg force measurements showed that this shift cannot be explained by the peak normal force of an isolated leg. Instead, force profiles reconstructed for the two phase-shifted tripod groups revealed that gait timing reshapes the temporal distribution of combined normal and shear forces. In doing so, it modulates both the effective propulsion window and the propulsive shear force acting within the propulsion window, explaining the observed differences in locomotion performance. 
Building on this mechanism, we developed a reduced-order predictive model that estimates phase-dependent granular resistive forces, identifies the effective propulsion window from the combined normal force, and predicts robot displacement by integrating the shear force over this window. The model reproduced the experimentally observed effects of surface inclination on locomotion performance and optimal slow-phase timing. These results provide a basis for terrain-aware gait optimization, with potential applications to locomotion in challenging environments, such as search and rescue, and  terrestrial and planetary surface exploration.

\bibliographystyle{./bibliography/IEEEtran.bst}
\bibliography{./bibliography/terradynamics.bib, ./bibliography/mudforce.bib, ./bibliography/mudsensing.bib, bibliography/qian, bibliography/rolling, bibliography/slope.bib}


\clearpage

\end{document}